\documentclass[11pt,a4paper,onecolumn]{article}
\usepackage[a4paper, margin=1in]{geometry}
 \pdfoutput=1 
\usepackage{authblk}
\usepackage{amsmath,amssymb,graphicx,cite}

\usepackage{caption}
\usepackage{subcaption}

\usepackage{balance}

\usepackage{psfrag}
\usepackage{transparent}
\usepackage[lined,linesnumbered,ruled]{algorithm2e}

\usepackage{color}  

\newcommand{\cb}[1]{{\boldsymbol{#1}}}
\newcommand{\cp}[1]{\ifmmode {\mathcal{#1}}\else ${\mathcal{#1}}$\fi}
\newcommand{\balpha}{\boldsymbol{\alpha}}
\newcommand{\bbeta}{\boldsymbol{\beta}}
\newcommand{\bgamma}{\boldsymbol{\gamma}}

\newcommand{\bA}{\boldsymbol{A}}

\newcommand{\bI}{\boldsymbol{I}}
\newcommand{\bK}{\boldsymbol{K}}

\newcommand{\bR}{\boldsymbol{R}}

\newcommand{\bm}{\boldsymbol{m}}
\newcommand{\bn}{\boldsymbol{n}}

\newcommand{\br}{\boldsymbol{r}}

\newcommand{\psh}[2]{\langle{#1},{#2}\rangle_{\cp{H}}}

\newcommand{\mr}[1]{{\bm_{\lambda_{#1}}}}

\newcommand{\MM}{{\boldsymbol{M}}}

\newcommand{\KK}{\boldsymbol{K}}

\providecommand{\keywords}[1]{\textbf{\textit{Index terms---}} #1}

\title{Band selection in RKHS for fast nonlinear unmixing \\of hyperspectral images}
 \author{T. Imbiriba$^{\,\star}$ \quad J. C. M. Bermudez$^{\,\star}$ \quad C. Richard$^{\,\dagger}$
\quad J.-Y. Tourneret$^{\,\ddagger}$
\thanks{This work was partly supported by CNPq under grants Nos 307071/2013-8, 400566/2013-3, 474735/2012-5 and 141094/2012-5, and by the Agence Nationale pour la Recherche, France, (Hypanema project, ANR-12- BS03-003), and by ANR-11-LABX-0040-CIMI within the program ANR-11-IDEX-0002-02.}}

\affil{$^{\star}$ Federal University of Santa Catarina, Florian\'opolis, SC, Brazil \\
    $^{\dagger}$Universit\'e de Nice Sophia-Antipolis, CNRS, Nice, France\\
    $^{\ddagger}$University of Toulouse, IRIT-ENSEEIHT, CNRS, Toulouse, France
    }

\begin{document}
%
\maketitle
\begin{abstract}
The profusion of spectral bands generated by the acquisition process of hyperspectral images generally leads to high computational costs. Such difficulties arise in particular with nonlinear unmixing methods, which are naturally more complex than linear ones. This complexity, associated with the high redundancy of information within the complete set of bands, make the search of band selection algorithms relevant. With this work, we propose a band selection strategy in reproducing kernel Hilbert spaces that allows to drastically reduce the processing time required by nonlinear unmixing techniques. Simulation results show a complexity reduction of two orders of magnitude without compromising unmixing performance. 
\end{abstract}
%
%
\section{Introduction}
\label{sec:intro}

 
 Hyperspectral images (HI) consist of hundreds or even thousands of contiguous spectral samples ranging from the visible to the near infrared portions of the light spectrum. HIs trade spacial for spectral resolution~\cite{Landgrebe:2002p5665}, a consequence of which is that  observed HI pixels can be mixtures of the spectral signatures of the materials present in the scene. The spectral signature of each material is usually available as a vector whose elements are proportional to the reflectances associated with that material at each frequency band. Such vectors are typically called endmembers due to their geometrical interpretation in the linear mixing case.  The analysis of HIs frequently aims at unmixing the spectral information present at each pixel in the image, a study known as hyperspectral unmixing (HU). Unmixing problems can be cast as supervised or unsupervised learning problems depending on whether the endmembers are known or not. 
 
Several models have been proposed in the literature to describe the mixing process of spectral information in HIs.  The simplest one is the linear mixing model (LMM), in which each observed pixel spectrum is modeled as a linear combination of the endmembers~\cite{Bioucas-Dias-2013-ID307}.  Though the LMM simplifies the mathematical treatment of the unmixing problem, it has been recognized that significant nonlinear effects are often present in the spectral mixing occurring in real images~\cite{Dobigeon-2014-ID322}.  Nonlinear mixing occurs, for instance, due to multiple interactions among light and different materials during the acquisition process.  Recognition of such nonlinear effects has led to several nonlinear mixing models for HI processing. These models include adding cross-terms of different endmembers to the LMM~\cite{Boardman1993}, using bilinear mixture models~\cite{halimi2011}, post-nonlinear mixing models~\cite{altmann2011:icassp,chen2013estimating}, and kernel-based models~\cite{Chen-2013-ID321}. Most nonlinear unmixing techniques are based on Bayesian inference~\cite{halimi2011}, on using manifold learning techniques and geodesic distances~\cite{nguyen2012hyperspectral,honeine2009solving,honeine2011preimage}, or processing data in reproducing kernel Hilbert spaces (RKHS)~\cite{Chen-2013-ID321,chen2013nonlinear2}.
 
 
One of the problems in practical implementation of unmixing algorithms is the profusion of spectral bands generated in the acquisition process, which leads to high computational costs.  This is especially true for nonlinear unmixing algorithms, which are naturally more complex than linear techniques.  Such inherent complexity, associated with the high redundancy within the complete set of bands, make the search of band selection techniques natural and relevant~\cite{chang2014hyperspectral}. Several band selection algorithms have been proposed for linearly mixed HIs, which generally requires solving an optimization problem~\cite{chang2014progressive}. Nonlinear unmixing presents an even more challenging problem for band selection. 

In this paper, we propose a technique for band selection in nonlinear supervised HI unmixing problems.  This method applies a kernel k-means algorithm to identify nonlinearly separable clusters of spectral bands in the corresponding RKHS. The solution is evaluated using the SK-Hype nonlinear unmixing algorithm~\cite{Chen-2013-ID321} on the selected bands. Simulation results indicate a complexity reduction of two orders of magnitude without compromising unmixing performance.

The paper is organized as follows. First, we state the unmixing problem and introduce usual nonlinear mixing models. Then, we describe a nonlinear unmixing algorithm operating in RKHS. Next, we introduce our band selection algorithm based on kernel k-means. We provide promising simulation results to illustrate the performance of our approach. Finally, we present some concluding remarks.

\section{Mixture Models}\label{sec:MM}

Each observed HI pixel can be written as a function of the endmembers and a noise component that covers the unknown or unmodeled factors in the system:
\begin{equation}
       \label{eq:model}
        \br = {\bf\Psi}(\MM) + \cb{n}
\end{equation}
with $\br =[r_1,\ldots, r_L]^\top$ the observed pixel vector over $L$ spectral bands, $\MM = [\bm_1,\ldots,\bm_R]$ the $L\times R$ matrix of endmembers $\bm_i$, $\bn$ a noise vector, and $\cb{\Psi}$ an unknown function. Several mixing models have been presented in the literature, depending on the linearity or nonlinearity of $\cb{\Psi}$, type of mixture, scale, and other properties~\cite{Dobigeon-2014-ID322}. 

\subsection{The Linear Mixing Model (LMM)}

The LMM assumes that $\cb{\Psi}$ is a convex combination of the endmembers, that is,
\begin{equation}
\begin{split}
\br &= \MM \balpha + \bn\\
 \text{subject to }&\,\cb{1}^\top\balpha = 1, \,\, \text{and}\,\, \balpha \succeq 0
 \label{eq:linForm}
\end{split}
\end{equation}
 where the vector $\balpha = [\alpha_1,\ldots,\alpha_{R}]^\top$ contains the proportions (or abundances) of each endmember in $\MM$ and, therefore, cannot be negative and should sum to one. The observation $r_\ell$ at the $\ell$-th spectral band in~\eqref{eq:linForm} can be written as 
 \begin{equation}
  r_\ell = \bm_{\lambda_\ell}^\top\balpha + n_\ell
  \label{eq:LMM_band}
 \end{equation}
where $\bm_{\lambda_\ell}^\top$ is the $\ell$-th row of $\MM$. For $n_\ell = 0$ (noiseless case), the sum-to-one and positivity constraints on the abundances in~\eqref{eq:linForm} confine the data (pixels) to a simplex, for which the vertices are the endmembers. 


\subsection{Nonlinear Mixing Models} \label{sec:NonlinearMixingModels}

Different parametric nonlinear models have been proposed in the literature (see, e.g.,~\cite{Dobigeon-2014-ID322} and references therein). Here we review two popular models that will be used later on to generate synthetic data for evaluation purposes.

The \emph{generalized bilinear model} (GBM)~\cite{halimi2011} is given by
\begin{equation}
\begin{gathered}
\br = \MM\balpha + \sum_{i=1}^{R-1}\sum_{j=i+1}^{R}\delta_{ij}\alpha_i\alpha_j\bm_i\odot\bm_j + \bn \\
 \text{subject to }\,\cb{1}^\top\balpha = 1, \,\, \text{and}\,\, \balpha \succeq 0
\end{gathered}
 \label{eq:GBM_orig}
\end{equation}
where the parameters $\delta_{ij}\in[0,1]$ govern the amount of nonlinear contribution, and $\odot$ denotes the Hadamard product.  For simplicity, we consider in the following a simplified version of this model, with a single parameter $\delta$ to control the nonlinear contribution such that $\delta_{ij}=\delta$ for all $(i,j)$.

The \emph{post nonlinear mixing model} (PNMM)~\cite{Jutten2003} is
\begin{equation}
 \br = \cb{g}(\MM\balpha) + \bn
\end{equation}
where $\cb{g}(\cdot)$ is a nonlinear function applied to the result of a linear mixing. The PNMM can represent a wide range of nonlinear mixing models, depending on $\cb{g}(\cdot)$. For instance, the PNMM considered in~\cite{Chen-2013-ID321}  is defined as
\begin{equation}
 \br = (\MM\balpha)^\xi + \bn
 \label{eq:pnmm_chen}
\end{equation}
where $(\cb{v})^{\xi}$ denotes the exponential value $\xi$ applied to each entry of the input vector $\cb{v}$.  The PNMM was explored in other works using different forms for $\cb{g}(\cdot)$ applied to HU~\cite{Altmann-2013-ID308, altmann2011tr, altmann2011:icassp}. 

The GBM~\eqref{eq:GBM_orig} and the PNMM~\eqref{eq:pnmm_chen} nonlinear mixing models consider mainly the scattering effect in light interactions with endmembers. Other models consider different nonlinear effects, depending on the characteristics of the application, such as the type of mixture, the types of materials, the geometry of the reflection surface, and the constraints over the abundances~\cite{fan2009,HapkeBook1993,Borel:1994tp,Somers:2009p6577,  Ray:1996tp}. More importantly, these informations are usually missing in HU problems. Therefore, it makes sense to consider nonlinear unmixing strategies that do not make strong assumptions about the nonlinearity in the mixture.

\section{Nonlinear unmixing strategy}
\label{sec:ls-svm}

This section reviews the SK-Hype algorithm\footnote{Matlab code available at www.cedric-richard.fr} for nonlinear unmixing of HIs~\cite{Chen-2013-ID321}. It considers the mixing model consisting of a linear trend parametrized by the abundance vector $\balpha$ and a nonlinear residual component $\psi$.  This model is given by
\begin{equation}
\label{eq:nlmodel}
r_\ell = u\,\balpha^\top \bm_{\lambda_\ell} + (1-u)\,\psi(\bm_{\lambda_\ell}) + n_\ell
\end{equation}
where $u\in [0,1]$ controls the amount of linear contribution to the model and $\psi(\cdot)$ is an unknown function in an RKHS $\cp H$. 

Kernel methods are efficient machine learning techniques that were initially introduced for solving nonlinear classification and regression problems. They consist of linear algorithms operating in high dimensional feature spaces into which the data have been mapped using kernel functions~\cite{Vapnik1995}. These approaches are based on the framework of reproducing kernels which states that, for any positive kernel $\kappa(\mr{i},\mr{j})$, there exists a Hilbert space $\cp H$ with inner product $\psh{\cdot\,}{\cdot}$ and a mapping
\begin{eqnarray}
	{\bf\Phi}:\quad {\mathbb R}^L&\longrightarrow & H \\
	 \mr{i} &\longmapsto& \kappa(\cdot,\mr{i})
\end{eqnarray}
such that $\kappa(\mr{i},\mr{j}) = \psh{\bf\Phi(\mr{i})}{\bf\Phi(\mr{j})}$. This last property allows to implicitly compute inner products in $\cp H$ by evaluating a real function, $\kappa(\mr{i},\mr{j})$, in the input space. Other useful properties are $\psi(\bm_{\lambda_j})=\langle\psi,\kappa(\cdot,\bm_{\lambda_j})\rangle_\cp{H}$ for all $\psi$ in $\cp{H}$, and the reproducing property 
$\kappa(\bm_{\lambda_i},\bm_{\lambda_j})=\langle\kappa(\cdot,\bm_{\lambda_i}),\kappa(\cdot,\bm_{\lambda_j})\rangle_\cp{H}$.

The optimization problem proposed in~\cite{Chen-2013-ID321} for estimating the unknown variables $\balpha$, $\psi(\cdot)$ and $u$ in \eqref{eq:nlmodel} is
 \begin{equation}
 \label{eq:optproblem}
 \begin{split}
\min_{\balpha,\psi,u} \frac{1}{2}&\left(\frac{1}{u}\|\balpha\|^2+\frac{1}{1-u}\|\psi\|_\cp{H}^2\right)\\
&+\frac{1}{2\mu}\sum_{\ell=1}^L (r_\ell-\balpha^\top\mr{\ell}-\psi(\mr{\ell}))^2
\end{split}
 \end{equation}
subject to $\balpha \succeq \bf{0}$ and $\cb{1}^\top\balpha = 1$. This convex problem can be solved using a two stage alternating iterative optimization procedure with respect to $(\balpha,\psi)$ and $u$.

\subsection{Solving with respect to $(\balpha, \psi)$}

Introducing the Lagrange multipliers $\bbeta$ and $\bgamma$, the dual problem of~\eqref{eq:optproblem} for fixed $u$ is given by~\cite{Chen-2013-ID321} 
\begin{equation}
	\label{eq:dual.algo2}
	\begin{split}
      		\max_{\bbeta,\bgamma}\,\, &G(u,\bbeta,\bgamma)  =\\ 
		&-\frac{1}{2}\left(
			\begin{array}{c}
				\bbeta \\ \hline
 				\bgamma
			\end{array}\right)^{\!\!\!\top}
			\left(
			\begin{array}{c|c}
				\bK_{u}+\mu\bI  & u\MM \\ \hline
 				u\MM^\top & u\bI
			\end{array}\right)
			\left(
			\begin{array}{c}
				\bbeta \\ \hline
 				\bgamma
			\end{array}\right)\\
			&+ \left(
			\begin{array}{c}
				\br \\ \hline
 				\bf{0}
			\end{array}\right)^{\!\!\!\top}
			\left(
			\begin{array}{c}
				\bbeta \\ \hline
 				\bgamma
			\end{array}\right)
			\\
      		& \text{subject to} \quad \bgamma \succeq \bf{0}
	\end{split}
\end{equation}
with $\bK_{u} = u\MM\MM^\top + (1-u)\KK$, and $\KK$ the Gram matrix such that  $\KK_{ij}=\kappa(\mr{i},\mr{j})$. This leads to the following solution of primal problem:
\begin{equation}
	\label{eq:solution.algo2}
	\left\{
    		\begin{array}{ll}
    			\balpha^\star = \frac{\MM^\top\bbeta^\star+\bgamma^\star}{\cb{1}^\top(\MM^\top\bbeta^\star+\bgamma^\star)} \\
    			\psi^\star = (1-u)\sum_{\ell=1}^L \beta_\ell^\star\,\kappa(\cdot,\mr{\ell}) \\					 
        			e_\ell^\star = \mu\,\beta_\ell^\star
    	\end{array}
	\right.
\end{equation}
where $\bbeta^\star$ and $\bgamma^\star$ are the solutions of \eqref{eq:dual.algo2}. 


\subsection{Solving with respect to $u$}
Using the stationary conditions~\eqref{eq:solution.algo2}, the optimum value for $u$ given $\bbeta^\star$ and $\bgamma^\star$ can be computed each iteration as~\cite{Chen-2013-ID317}
\begin{equation}
	{ u^\star = \left( 1+ (1-u^\star_{-1})\sqrt{\frac{{\bbeta^\star}^\top\bK\bbeta^\star}{\|\MM^\top\bbeta^\star+\bgamma^\star \|^2}}\right)}
\end{equation}
where $u^\star_{-1}$ is the optimum $u^\star$ obtained at the previous iteration.

\section{Band selection and unmixing}
\label{sec:bandSelection}

Band selection is the strategy of choice for reducing the complexity of HI processing when the original spectral information of each pixel needs to be preserved~\cite{Chang:2006jg}. Most existing band selection algorithms have been derived assuming linearly mixed HIs.  To the best of our knowledge, the problem of band selection for a single nonlinearly mixed pixel and with the objective of reducing the complexity of the unmixing process is still largely untreated in the literature. As each pixel may be better characterized by a different set of bands, we propose a simple supervised band selection approach to be applied to individual pixels so that only those endmembers present in each pixel, that is, the matrix $\MM$ for that pixel, are considered.  The clustering technique proposed in the following is based on the \emph{kernel k-means} (KKM) algorithm~\cite{Tzortzis-2009-ID339}, and uses the fact that each band of the HI is a function of the elements in one row of matrix $\MM$. The choice of the KKM algorithm is due to the practical assumption that we lack information about the nonlinearity associated to the endmember mixing in each HI. 

\subsection{Kernel k-means for band-selection}

For band selection we consider each row of $\MM$ as an element of a vector space, and search for a set of $K$ disjoint clusters $\cp{C}_1,\ldots,\cp{C}_K$ in that space. Then, a unique wavelength is chosen using the KKM algorithm to represent each cluster.  KKM is an extension of the standard k-means clustering algorithm that identifies nonlinearly separable clusters~\cite{Tzortzis-2009-ID339}. It maps the data implicitly to a RKHS $\cp{H}$ where it performs a conventional k-means algorithm. Since the computation of the centroids in space $\cp{H}$ is intractable, KKM algorithms use the reproducing property~\cite{Aronszajn1950} to directly compute squared distances between points in a cluster $\cp{C}_k$. Therefore, given a cluster $\cp{C}_k$ enclosing points $\{\kappa(\cdot,\mr{\ell})\}_{\ell\in \cp{C}_k}$, we can write the centroid $\mu_k$ as
\begin{equation}
  \mu_k = \frac{1}{N_k} \sum_{i\in\cp{C}_k}\kappa(\cdot, \mr{i})
\end{equation}
where $N_k$ is the number of points in $\cp{C}_k$. The squared distances to the centroid of $\cp{C}_k$ are then computed as
\begin{equation}
\begin{split}
 \|\kappa(\cdot,\mr{\ell}) - \mu_k\|^2_{\cp H} &= \kappa(\mr{\ell},\mr{\ell})\\
 &- \frac{1}{N_k} \sum_{i\in \cp{C}_k}\kappa(\mr{\ell},\mr{i})\\
 &+ \frac{1}{N_k^2}\sum_{i\in \cp{C}_k}\sum_{j\in C_k}\kappa(\mr{i},\mr{j})
 \end{split}
\end{equation}
and the cluster error is defined as
\begin{equation}
 E(\mu_1,\ldots,\mu_K) = \sum_{k=1}^{K}\sum_{\ell\in \cp{C}_k} \| \kappa(\cdot,\mr{\ell}) - \mu_k\|^2_{\cp H}.
\end{equation}

To preserve the original HI band information, we propose to represent cluster $\cp{C}_k$ by the band $\ell_k$ corresponding to the closest point to the centroid $\mu_k$: 
\begin{equation}
 \ell_k =  \mathop{\arg\min}_{\ell\in\cp{C}_k} \|\kappa(\cdot,\mr{\ell}) - \mu_k\|^2_{\cp{H}}.
 \label{eq:bandSelection}
\end{equation}

The \emph{global kernel k-means} (GKKM) algorithm uses the principles above for incremental clustering~\cite{Tzortzis-2009-ID339}. GKKM  avoids poor local minima and produces near-optimal solutions that are robust to cluster initialization. A fast GKKM (FGKKM) version that performs a unique KKM run and greatly reduces the complexity of the algorithm can also be used. Algorithm~\ref{alg:FKKMBS} details the application of FGKKM algorithm using \eqref{eq:bandSelection} for band selection. We refer to Algorithm~\ref{alg:FKKMBS} as KKMBS for short.

\begin{algorithm}
\SetKwInOut{Input}{Input}
\SetKwInOut{Output}{Output}
\caption{FGKKM Band Selection (KKMBS)~\label{alg:FKKMBS}}
 \Input{The $L\times R$ matrix $\MM$ and the desired number of bands $N_b$.}
 \Output{Selected band indices $\cb{\ell}$.}
 
 \% Find clusters \\
$[\cp{C}_1,\ldots, \cp{C}_{N_b}] = \text{FGKKM}(\MM,N_b)$; \\
\% Find the lines of $\MM$ closer to the centroids in $\cp{C}_1,\ldots, \cp{C}_{N_b}$ \\
  \For{$k=1$ \KwTo $N_b$}{ 
    $\ell_k = \mathop{\arg\min}_{\ell\in \cp{C}_k} \|\kappa(\cdot,\mr{\ell}) - \mu_k\|^2_{\cp{H}}$\;
}
 \KwRet $\cb{\ell}$\
\end{algorithm}

\subsection{Unmixing of reduced data}
\begin{figure}[htb]
 \centering 
\def\svgwidth{200pt}
\hspace{1cm}
 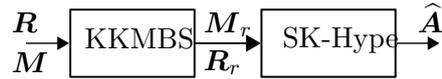
 \caption{Band selection and unmixing.}
 \label{fig:blockDiag}
\end{figure}

For a given $N$-pixel HI region with a known endmember matrix $\MM$, define $\bR=[\br_1,\ldots,\br_N]$ as the matrix of observations. The unmixing problem then reduces to an inversion step for which we solve the optimization problem described in Section~\ref{sec:ls-svm} to obtain the estimated abundances given by $\bA=[\balpha_1^\star,\ldots,\balpha_N^\star]$. We propose to replace matrices $\bR$ and $\MM$ by their reduced versions $\bR_r$ and $\MM_r$ containing only $N_b$ selected bands obtained with Algorithm~\ref{alg:FKKMBS}. Then, the SK-Hype algorithm can be used to perform data unmixing and to estimate the abundance matrix as shown in Figure~\ref{fig:blockDiag}.

\section{Simulation results}
\label{sec:simulations}

This section presents simulation results with synthetic data to illustrate the potential of the band selection technique. The HI was built using measured spectra from eight materials: alunite, calcite, epidote, kaolinite, buddingtonite, almandine, jarosite and lepidolite. Their spectra were selected from the spectral library of the {ENVI} software and consisted of 420 contiguous bands, covering wavelengths from 0.3951 to 2.56 micrometers. We compared the results obtained with 2 nonlinear mixing models using 5 unmixing algorithm implementations and 2 performance measures: the \emph{root mean square error} (RMSE) and the \emph{relative execution time} (RET). 
\begin{equation}
	\text{RMSE}=\|\bA-\widehat{\!\bA}\|_{F}/(NR)
\end{equation}
where $\|\cb{X}\|_F$ denotes the Frobenius norm of matrix $\cb{X}$. The RET for the $p$-th algorithm was determined as $\text{RET}_{p} = t_p/t_{\rm{FCLS}}$, where $t_p$ is the execution time for the $p$-th algorithm and $t_{\rm{FCLS}}$ is the execution time for the Fully Constrained Least Squares (FCLS) algorithm. 


The data were nonlinearly mixed using the PNMM and GBM models presented in Section~\ref{sec:NonlinearMixingModels}. For PNMM, we used $\xi = 0.7$, and for GBM, we considered $\delta = 1$. We generated 6 databases using the two models and different number of endmembers (5 and 8). Each database had 2000 pixels generated using random abundances drawn uniformly from the simplex defined by~\eqref{eq:linForm}. The additive noise power $\sigma_n^2$ was chosen to produce a 21dB SNR. For each dataset we unmixed the data using FCLS, SK-Hype without band selection and the proposed algorithm for $N_b=[10, 100, 300]$. For both SK-Hype and KKMBS, we used the Gaussian kernel~\cite{Scholkopf1999} with bandwidth $\sigma_k^2 = 0.3$. Table~\ref{tab:pnmm} presents the RMSEs and the RETs obtained for each simulation. The RETs within parentheses include the time for both band separation and unmixing as indicated.  These results show comparable RMSEs for all nonlinear unmixing results and very significant improvements in RET when using the proposed band selection approach (about 145 times for $N_b=10)$.  

\begin{figure*}[t]
\centering 
\begin{subfigure}[b]{0.45\textwidth}
 \includegraphics[width=\textwidth]{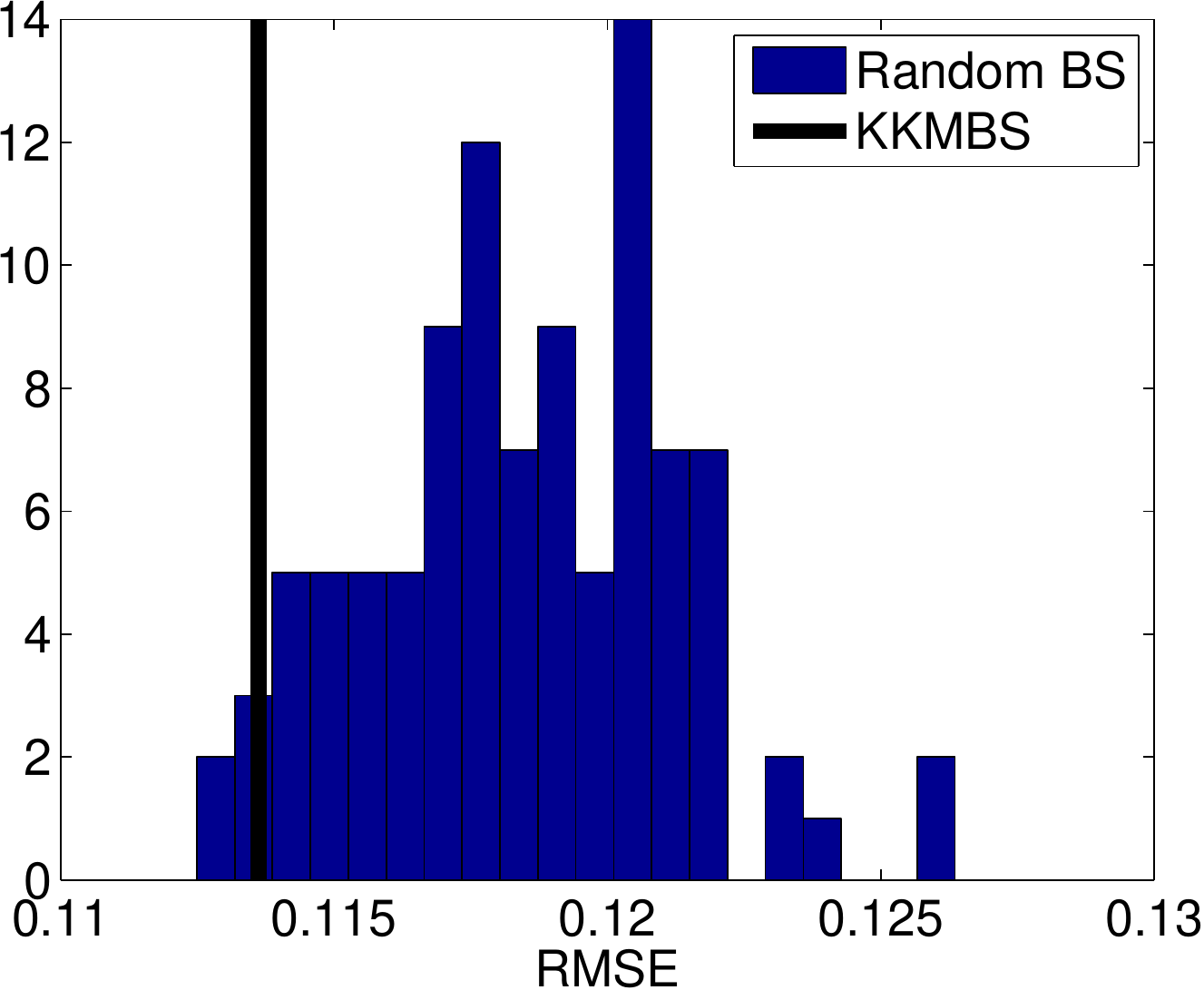}
 \caption{10 bands.}
 \end{subfigure} \qquad
\begin{subfigure}[b]{0.45\textwidth}
 \includegraphics[width=\textwidth]{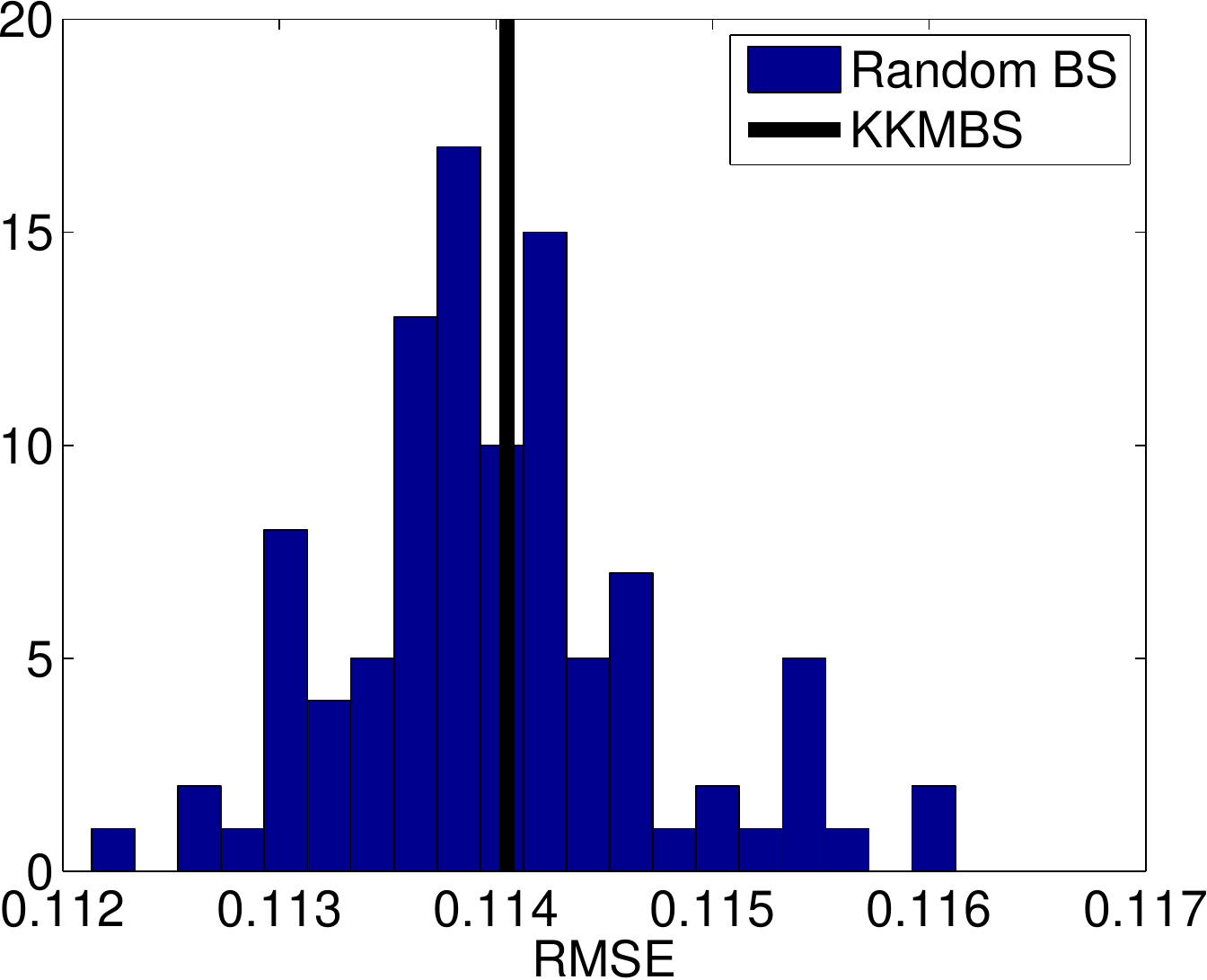}
 \caption{100 bands.}
 \end{subfigure}
 \caption{RMSE for 10 and 100 bands.}\label{fig:randHists}
\end{figure*}

Table~\ref{tab:changeNonlin} shows the results when the nonlinearity parameters ($\delta$ for GBM and $\xi$ for PNMM) were set differently for each band. Parameter $\delta$ ($\xi$) varied in the interval $[0.5,1]$ ([0.5, 0.9]) with steps of 0.05 (0.04), changing at every 42 bands. These results corroborate the results shown in Table~\ref{tab:pnmm}. We have also compared the results of KKMBS with those obtained after randomly selecting the bands. Figure~\ref{fig:randHists} shows the histograms of the RMSE when selecting 10 and 100 bands. It is clear that the KKMBS leads to significantly better results when few bands are selected.

\begin{table}[htb]
\footnotesize
\caption{RMSE and RET for SNR $=21$dB and random abundances.}\label{tab:pnmm}
 \begin{center}
\begin{tabular}{|l|l|l|l|l|}
\cline{2-5}
\multicolumn{1}{c}{}&\multicolumn{4}{|c|}{PNMM} \\ \hline
& \multicolumn{2}{|c|}{5 endmembers} & \multicolumn{2}{|c|}{8 endmembers}\\ \hline
Alg. ($N_b$) & RMSE & RET (BS+HU) & RMSE & RET (BS+HU)\\ \hline
FCLS & 0.1893 & 1 & 0.1243  & 1 \\ \hline 
SK-Hype & 0.1136 & 2690.6 & \textbf{0.0762} & 3028.8  \\ \hline
SK-Hype(10) & \textbf{0.1114} & 16.0 (18.1) &  0.0775 & 16.8 (18.7) \\ \hline
SK-Hype(100) & 0.1150  & 107.7 (129.1) &  0.0766 & 118.6 (144.8) \\ \hline  
SK-Hype(300) & 0.1139  & 1226.7 (1327.1) &  0.0763 & 1331.5 (1452.7) \\ \hline
& \multicolumn{4}{|c|}{GBM} \\ \hline
FCLS & 0.2419 & 1 &  0.1836 & 1 \\ \hline 
SK-Hype & 0.1080  & 3320.7 & 0.0738 & 3072.6 \\ \hline
SK-Hype(10) &  \textbf{0.1037} & 24.1 (26.9) & \textbf{0.0712} & 18.7 (21.1) \\ \hline
SK-Hype(100) & 0.1095 & 157.0 (194.6) & 0.0741 & 119.7 (148.1) \\ \hline  
SK-Hype(300) & 0.1083 & 1548.6 (1676.0)  & 0.0738 & 1475.2 (1597.7) \\ \hline
\end{tabular}
\end{center}
\end{table}

\begin{table}[htb]
\footnotesize
\caption{RMSE and RET for different nonlinearities in each band.}\label{tab:changeNonlin}
 \begin{center}
\begin{tabular}{|l|l|l|l|l|}
\cline{2-5}
\multicolumn{1}{c}{}&\multicolumn{4}{|c|}{PNMM} \\ \cline{2-5}
\multicolumn{1}{c}{}&\multicolumn{2}{|c|}{5 Endmembers} & \multicolumn{2}{|c|}{8 Endmembers}\\ \hline
Alg. ($N_b$) & RMSE & RET (BS+HU) & RMSE & RET (BS+HU)\\ \hline
FCLS & 0.1966  & 1  & 0.1521  & 1  \\ \hline  
SK-Hype & \textbf{0.1127}  & 3744.9  & \textbf{0.0765}  & 3672.2  \\ \hline  
SK-Hype(10) & 0.1131  & 21.7 (24.3) & 0.0827  & 21.5 (24.4) \\ \hline  
SK-Hype(100) & 0.1140 & 153.4 (184.2) & 0.0771 & 140.3 (170.8) \\ \hline  
SK-Hype(300) & 0.1129 & 1764.3 (1907.5) & \textbf{0.0765} & 1601.0 (1743.7) \\ \hline   
&\multicolumn{4}{|c|}{GBM} \\ \hline
&\multicolumn{2}{|c|}{5 Endmembers} & \multicolumn{2}{|c|}{8 Endmembers}\\ \hline
FCLS & 0.0527  & 1  & 0.0390  & 1  \\ \hline  
SK-Hype & 0.1090  & 3432.5  & 0.0745  & 3825.1  \\ \hline  
SK-Hype(10) & \textbf{0.1053}  & 20.8 (23.6) & \textbf{0.0730} & 20.9 (23.0) \\ \hline  
SK-Hype(100) & 0.1105 & 142.2 (171.5) & 0.0748  & 142.9 (173.8) \\ \hline  
SK-Hype(300) & 0.1093 & 1612.8 (1735.2) & 0.0745  & 1637.8 (1778.1) \\ \hline  
\end{tabular}
\end{center}
\end{table}


\section{Conclusions}
\label{sec:conclusions}
This work proposed a supervised band selection strategy to reduce the complexity of nonlinear hyperspectral data unmixing without compromising the accuracy of abundance estimation.  Significant reduction in processing time was achieved in all cases tested. These results suggest the possibility of important complexity reduction for nonlinear HI processing algorithms without performance loss. It is conjectured that presented performance can be further improved by removing redundancy in the data through a more specialized band selection algorithm. This is the topic of a work in progress. 


%


\balance

\bibliographystyle{IEEEbib}
\bibliography{hyperspectral}

\balance

\end{document}